# COMPARING SPECTROSCOPY MEASUREMENTS IN THE PREDICTION OF IN VITRO DISSOLUTION PROFILE USING ARTIFICIAL NEURAL NETWORKS

Mohamed Azouz Mrad, Kristóf Csorba
Dorián László Galata, Zsombor Kristóf Nagy and Brigitta Nagy

Department of Automation and Applied Informatics,
Budapest University of Technology and Economics, Budapest, Hungary

## ABSTRACT

*Dissolution testing is part of the target product quality that is essential in approving new products in the pharmaceutical industry. The prediction of the dissolution profile based on spectroscopic data is an alternative to the current destructive and time-consuming method. Raman and near-infrared (NIR) spectroscopies are two fast and complementary methods that provide information on the tablets' physical and chemical properties and can help predict their dissolution profiles. This work aims to compare the information collected by these spectroscopy methods to support the decision of which measurements should be used so that the accuracy requirement of the industry is met. Artificial neural network models were created, in which the spectroscopy data and the measured compression curves were used as an input individually and in different combinations in order to estimate the dissolution profiles. Results showed that using only the NIR transmission method along with the compression force data or the Raman and NIR reflection methods, the dissolution profile was estimated within the acceptance limits of the f2 similarity factor. Adding further spectroscopy measurements increased the prediction accuracy.*

## 1. INTRODUCTION

In the pharmaceutical industry, a target product quality profile is a term used for the quality characteristics that a drug product should go through to satisfy the promised benefit from the usage and are essential in the approval of new products or the post-approval changes. A target product quality profile would include different essential characteristics. One of these is the in vitro (taking place outside of the body) dissolution profile [1]. A dissolution profile represents the concentration rate at which capsules and tablets emit drugs into the bloodstream over time. It is essential for tablets that yield a controlled release into the bloodstream over several hours. That offers many advantages over immediate release drugs, like reducing the side effects due to the reduced peak dosage and better therapeutic results due to the balanced drug release [2]. In vitro dissolution testing has been a subject of scientific research for several years and has become a vital tool for accessing product quality performance [3]. However, this method is destructive since it requires immersing the tablets in a solution simulating the human body. It is time-consuming as the measurements require taking samples over several hours. As a result, the tablets measured represent only a limited amount of the tablets produced, also called a batch. Therefore, there is a need to find different methods that do not have the limitations of the in vitro dissolution testing. The prediction of the dissolution profile based on spectroscopic data is an alternative on which many articles have been published and showed promising results. RAMAN and Near Infrared (NIR) spectroscopy are two evolving techniques in the pharmaceutical industry. The interaction of NIR and RAMAN with tablets in reflection (what is reflected from the tablet) and transmission (what is transmitted through the tablet) offer the opportunity to obtain information on the physical and chemical properties of the tablets that can help predict their dissolution profiles in few minutes without destroying them. RAMAN spectroscopy is very sensitive for analyzing Active Pharmaceutical Ingredient (APIs), the part of the drug that produces the intended effect. NIR spectroscopy, on the other hand, is better used for the tableting excipients, the substances added to aid in manufacturing the tablets. Hence, RAMAN and NIR are considered to be complementary methods, straight-forward, cost-effective alternatives, and non-destructive tools in the quality control process [4, 5]. The utilization of NIR and RAMAN spectroscopy in the pharmaceutical industry has been increasing quickly. They have been applied to determine content uniformity [6], detect counterfeit drugs [7], and monitor the polymorphic transformation of tablets [8].

RAMAN and NIR spectroscopies produce a large amount of data as they consist of measurements of hundreds of wavelengths. This data can be filtered out or maintained depending on how much valuable information it provides. The information can be extracted using multivariate data analysis techniques such as Principal Component Analysis (PCA). Several researchers have used spectroscopy data and multivariate data analysis techniques to predict the dissolution profiles. Zan-Nikos et al. worked on a model that permits hundreds of NIR wavelengths to be used to determine the dissolution rate [9]. Donoso et al. [10] used the NIR reflectance spectroscopy to measure the percentage of drug dissolution from a series of tablets compacted at different compressional forces using linear regression, nonlinear regression, and Partial Least Square (PLS) models. Freitas et al. [11] created a PLS calibration model to predict drug dissolution profiles at different time intervals and for media with different pH using NIR reflectance spectra. Hernandez et al. [12] used PCA to study the sources of variation in NIR spectra and a PLS-2 model to predict the dissolution of tablets subjected to different strain levels.

Artificial Neural Networks (ANNs) are suitable for complex and highly nonlinear problems. They have been used in the pharmaceutical industry in many aspects, such as the prediction of chemical kinetics [13], monitoring a pharmaceutical freeze-drying process [14], and solubility prediction of drugs [15]. ANN models have also been used to predict the dissolution profile based on spectroscopic data. Ebube et al. [16] trained an ANN model with the theoretical composition of the tablets to predict their dissolution profile. Galata et al. [17] developed a PLS model to predict the contained drotaverine (DR) and the hydroxypropyl methylcellulose (HPMC) content of the tablets, which are respectively the drug itself and a jellying material that slows down the dissolution, based on both RAMAN and NIR Spectra. They used the predicted values and the measured compression force as input to an ANN model to predict the dissolution profiles. Mrad et al. [18] used RAMAN and NIR spectroscopy along with the compression force to estimate the dissolution profiles of the tablets defined in 53-time points using ANN models. Using NIR spectra, RAMAN spectra, and the concentration force to predict the dissolution profile is a fast method that requires minimal human labor and makes it easier to evaluate a more significant amount of the batch. The decision of which measurements (Raman Reflection, Raman Transmission, NIR Reflection, NIR transmission, and compression force) to use in predicting dissolution profiles can be supported if the methods are compared. We aimed to support the decision of which measurements to use by comparing how well these measurements are in predicting dissolution profiles. In this paper, our goal was to extract helpful information from the NIR, RAMAN spectroscopy, and the compression curve of the tablets using a multivariate data analysis technique. ANN models were then created using the extracted information as input individually and in different combinations to predict the dissolution profiles represented in the 53-time points.

## 2. DATA AND METHODS

In Section 2, the data used will be described, and the methods used for the data pre-processing will be presented. The artificial neural network models will be presented, and finally, the error measurement methods adopted to evaluate the results.

### 2.1. Data Description

The chemical engineers in this paper prepared the NIR and RAMAN spectroscopy measurements (Figure. 1), along with the pressure curves extracted during the compression of the tablets. The data consists of the NIR reflection and transmission, RAMAN reflection and transmission spectra, the compression force-time curve, and the dissolution profile of 148 tablets (Figure. 2).

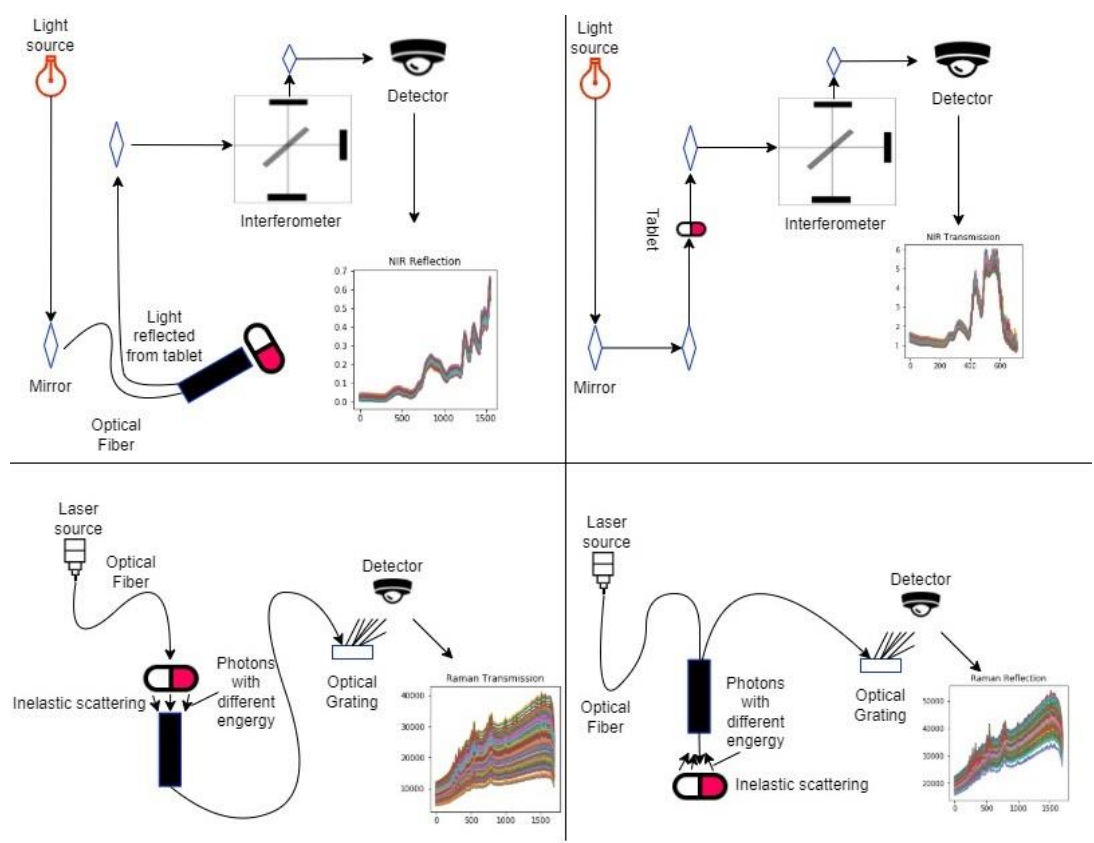

Figure 1. Spectroscopy methods: NIR reflection, NIR transmission, Raman transmission and Raman Reflection (Clockwise starting from Top left corner)

The tablets were produced with a total of 37 different settings. Three parameters were varied: drotaverine content, HPMC content, and the compression force. From each setting, four tablets were selected for analysis (37*4). The NIR and RAMAN measurements on the tablets were carried out by Bruker Optics MPA FT-NIR spectrometer, and Kaiser RAMAN RXN2 Hybrid Analyzer equipped Pharmaceutical Area Testing (PhAT) probe. The spectral range for NIR reflection spectra was 4,000–10,000 $cm^{-1}$ with a resolution of 8 $cm^{-1}$, representing 1556 wavelength points. NIR transmission spectra were collected in the 4000-15,000 cm$^{-1}$ wavenumber range with 32 cm$^{-1}$ spectral resolution representing 714 wavelength points. RAMAN spectra were recorded in the range of 200-1890 cm$^{-1}$ with 4 cm$^{-1}$ spectral resolution for transmission and reflection measurements representing 1691 points. Two spectra were recorded for each tablet in both NIR and RAMAN. The pressure during the compression of the tablet was recorded in 6037 time points. The dissolution profiles of the tablets were recorded using Hanson SR8-Plus in vitro dissolution tester. The length of the dissolution run was 24 hours. During this period, samples were taken at 53-time points (at 2, 5, 10, 15, 30, 45, and 60 min, after that once every 30 min until 1440 min).

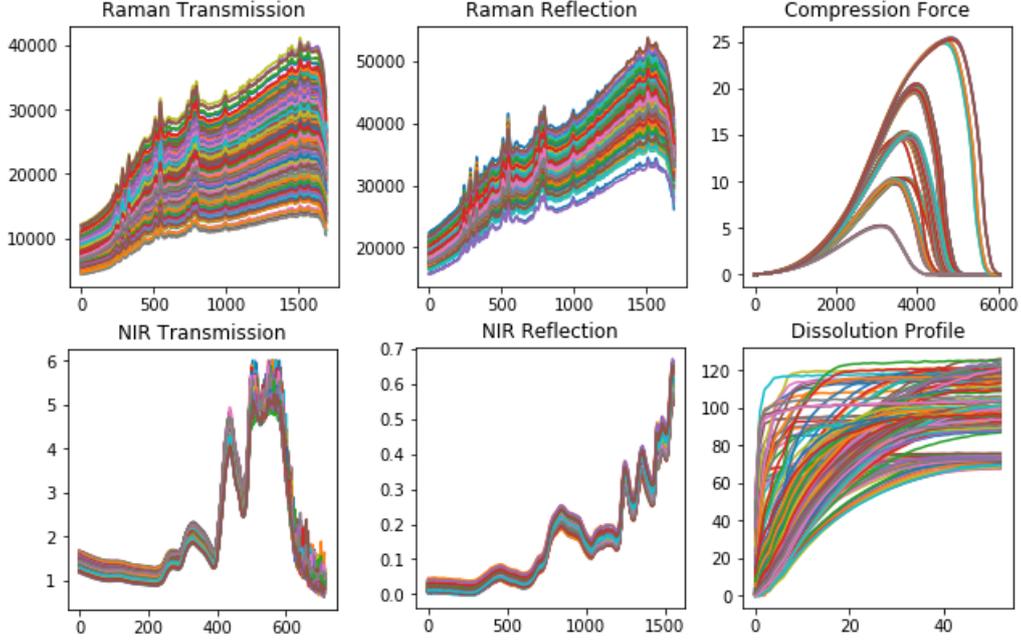

Figure 2. Dataset composed of NIR, RAMAN transmission and reflection the compression curve and the dissolution profiles

## 2.2. Data Analysis

The collected data were visualized and analyzed using MATLAB and Excel in order to detect and fix missed and wrong values: Setting first point of the dissolution curves to zero, detecting missed values, and fixing negative values found due to error of calibration, etc. Specifically, the data is represented in matrices $N_i^n$ for NIR transmission data and $M_j^n$ for NIR reflection data, where i= 1556, j=714. $R_k^n$ and $Q_k^n$ respectively for Raman reflection and transmission data where k=1691. $C_l^n$ for the compression force data where l=6037 and $P_s^n$ for the dissolution profiles where s=54. With n representing the number of samples which is equal to 296. All the different NIR, RAMAN and the compression force matrices have been standardized using scikit-learn preprocessing method: StandardScaler. StandardScaler fits the data by computing the mean and standard deviation and then centers the data following the equation $\text{Std}(NS) = (NS - u)/s$, where NS is the non-standardized data, u is the mean of the data to be standardized, and s is the standard deviation. All the different standardized NIR, RAMAN and the compression force matrices have been row-wise concatenated to form a new matrix $D_m^n$ where n=296 and m=i+j+2k+l=11686 as follow: $D_m^n = ( N_i^n | M_j^n | R_k^n | Q_k^n | C_l^n )$.

After standardization, PCA was applied to the different standardized matrices as well as the merged data $D_m^n$ in order to reduce the dimension of the data while extracting and maintaining the most useful variations. Basically, taking $D_m^n$ as an example we construct a symmetric m*m dimensional covariance matrix Σ (where m=11686) that stores the pairwise covariances between the different features calculated as follow:

$$\sigma_{j,k} = \frac{1}{n}\sum_{i=1}^{n}(x_j^{(i)} - \mu_j)(x_k^{(i)} - \mu_k) \qquad (1)$$

With $\mu_j$ and $\mu_k$ are the sample means of features j and k. The eigenvectors of Σ represent the principal components, while the corresponding eigenvalues define their magnitude. The eigenvalues were sorted by decreasing magnitude in order to find the eigenpairs that contains most of the variances. Variance explained ratios represents the variances explained by every principal component (eigenvectors), it is the fraction of an eigenvalue $\lambda_j$ and the sum of all the eigenvalues. The following plot (Figure. 3) shows the variance explained rations and the cumulative sum of explained variances. It indicates that the first principal component alone accounts for 50% of the variance. The second component account for approximately 20% of the variance.

The plot indicates that the seven first principal components combined explain almost 96% of the variance in D. These components are used to create a projection matrix W which we can use to map D to a lower dimensional PCA subspace D' consisting of less features:

$$D = [d_1, d_2, d_3, \dots d_m] \in R^m \rightarrow D' = DW, W \in R^{m*v} \qquad (2)$$
$$D' = [d_1, d_2, d_3, \dots d_m], d \in R^m \qquad (3)$$

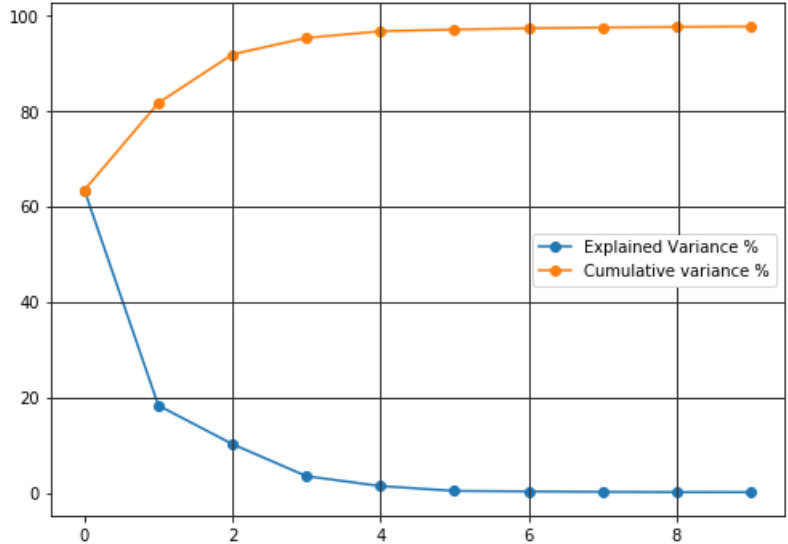

Figure 3. Explained PCA and Cumulative variances.

## 2.3. Artificial Neural Networks

ANN models were used to predict the dissolution profiles of the tablets. The models were created using the Python library Sklearn. Different ANN models were created, with different inputs and output targets each time. The models used the rectified linear unit activation function referred to as ReLU on the hidden layers and the weights on the models were optimized using LBFGS optimizer which is known to perform better and converge faster on dataset with small number of samples (296 in our case). Adam optimizer was tried as well but did not perform as good as LBGFS. The mean-squared error (MSE) was the loss function used by the optimizer in the different models. The training target for the models were the remaining part of the dissolution profiles, e.g., the dissolution curves are described in 53 points, if 10 points are used in the input then the remaining part of 43 points is the training target. The number of layers on the models and the number of neurons were optimized based on their performances. Regularization term has been varied in order to reduce overfitting. In each training, 16% of the training samples (49 samples) were selected randomly for testing. The accuracy of the model's predictions was calculated by evaluating the similarity of the predicted and measured parts of the dissolution profiles using the $f_2$ similarity values.

## 2.4. Error Measurement

Two mathematical methods are described in the literature to compare dissolution profiles [19]. A difference factor $f_1$ which is the sum of the absolute values of the vertical distances between the test and reference mean values at each dissolution time point, expressed as a percentage of the sum of the mean fractions released from the reference at each time point. This difference factor $f_1$ is zero when the mean profiles are identical and increases as the difference between the mean profiles increases:

$$f_1 = \sum_{t=1}^{n}|R_t - T_t|/\sum_{t=1}^{n}|R_t| * 100 \qquad (4)$$

Where $R_t$ and $T_t$ are the reference and test dissolution values at time $t$. The other mathematical method is the similarity function known as the $f_2$ measure, it performs a logarithmic transformation of the squared vertical distances between the measured and the predicted values at each time point. The value of $f_2$ is 100 when the test and reference mean profiles are identical and decreases as the similarity decreases.

$$f_2 = 50 \log_{10}\left[\left(1 + {1}/{n} \sum_{t=1}^{n}(R_t - T_t)^2\right)^{-0.5}\right) * 100 \qquad (5)$$

Values of $f_1$ between zero and 15 and of $f_2$ between 50 and 100 ensure the equivalence of the two dissolution profiles. The two methods are accepted by the FDA (U.S. Food and Drug Administration) for dissolution profiles comparison, however the $f_2$ equation is preferred, thus in this paper maximizing the $f_2$ will be prioritized.

## 3. RESULTS AND DISCUSSIONS

In this section the results after the PCA dimensionality reduction will be discussed. The results and the performance of the Artificial Neural Network models created will be presented.

### 3.1. Dimensionality reduction using PCA

Principal component analysis transformation was applied in a first step to the standardized NIR and Raman spectra recorded in reflection and transmission mode ($N_i^n, M_j^n, R_k^n, Q_k^n$ matrices) and the standardized compression force curve $C_l^n$, and in a second step on all the data merged in matrix $D_m^n$ in order to investigate the effect of the transformation on the merged and the separated data.

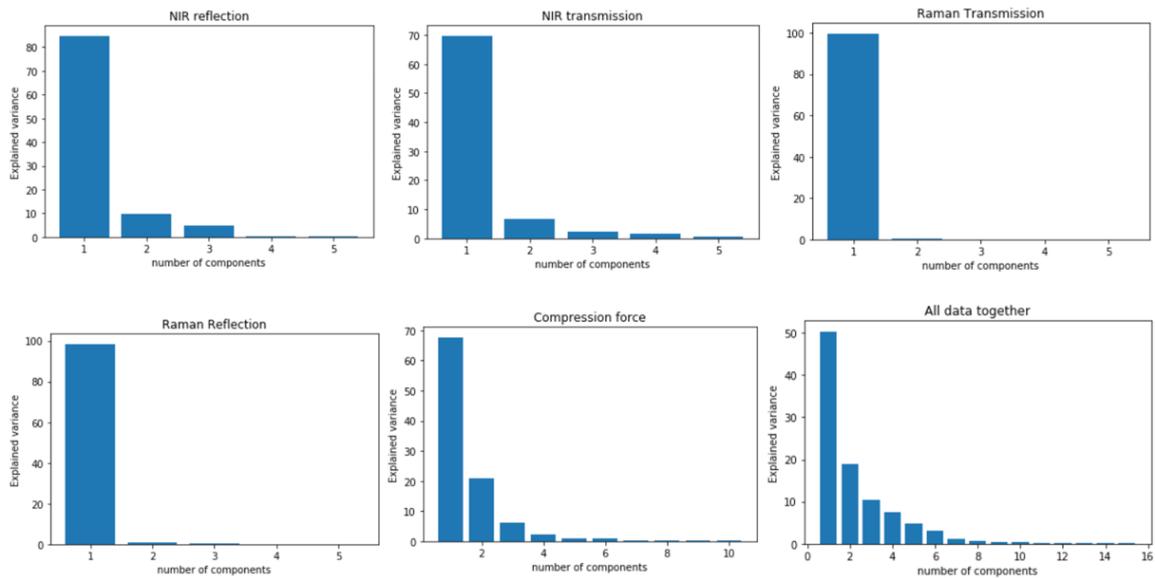

Figure 4. Explained variance of spectral data, compression force, and all data merged.

The resulting PCA decompositions, showed that in the case of NIR reflection, three principal components explaining 84.79%, 9.67% and 4.83% of the total variance in the data, respectively, leading to a cumulative explained variance of more than 99%. Four principal components explained more than 80% of the total variances of the NIR transmission data and 95% of the compression force data. However, for Raman transmission, the first principal component alone explains 99.69% of the variance in the data. The first two principal components explain 98.51% and 1.01% of the variance in the Raman Reflection data, respectively. For matrix $D_m^n$, 7 principal components explain more than 95 % of the variance and 33 explain more than 99% of the merged standardized data. These data resulting from the PCA decompositions were used as inputs for the Artificial neural network models individually and in different combinations in order to compare them based on how helpful they are in the prediction of dissolution profiles. For all measurements, the number of components explaining 99% of the total variance were kept.

## 3.2. Predicting the Dissolution Profile using Artificial neural network

The results showed that by using only one measurement as input, the artificial neural network models were not able to predict the dissolution profiles within the acceptance range (50-100) of the $f_2$ factor, as the maximum average $f_2$ was 47.56 using the compression force as input for the ANN model. Thus, further measurements were added in order to improve the results. By Combining two measurements, two ANN models were able to predict the dissolution profiles within the acceptance range of $f_2$. The first model used NIR transmission along with the compression force measurements as an input, this model was able to reach an $f_2$ average of 60.69. The second ANN model used the Raman Reflection with NIR Reflection methods, and had an $f_2$ average of 50.22. Further measurements were added to verify the effect on the prediction accuracy. The results showed that ANN models that used the combination of NIR Transmission and the compression force along with either Raman Reflection or NIR Reflection, were able to predict the dissolution profile with an $f_2 > 60$. The results show that NIR transmission and the compression force are very important in the prediction of dissolution profiles, adding further measurements to these two can slightly improve the results.

Table 1. Results of the predictions using one measurement.

|  | $F_2$ Results of the Prediction |
|---|---|
| RAMAN TRANSMISSION | 40.82 |
| RAMAN REFLECTION | 41.64 |
| NIR TRANSMISSION | 43.89 |
| NIR REFLECTION | 43.89 |
| COMPRESSION FORCE | **47.56** |

Table 2. Results of the predictions using combination of two measurements.

|  | $F_2$ Results of the Prediction |
|---|---|
| NIR TR+RAMAN RE | 45.55 |
| NIR TR+ Compression Force | **60.69** |
| NIR TR+NIR RE | 42.52 |
| NIR TR+RAMAN TR | 44.10 |
| RAMAN RE+ Comp Force | 47.73 |
| RAMAN RE + NIR RE | 50.22 |
| RAMAN RE+RAMAN TR | 43.05 |
| Comp Force+ NIR RE | 49.09 |
| Comp Force+ RAMAN TR | 47.68 |
| NIR RE+ RAMAN TR | 46.62 |

Table 3. Results of the predictions using combination of three measurements

|  | $F_2$ Results of the Prediction |
|---|---|
| NIR TR+RAMAN RE+ Comp | **61.03** |
| NIR TR+ RAMAN RE + NIR RE | 49.77 |
| NIR TR+RAMAN RE + RAMAN TR | 47.76 |
| NIR TR+ Comp+ NIR RE | **61.24** |
| NIR TR+ Comp+ RAMAN TR | 59.12 |
| NIR TR + NIR RE+ RAMAN TR | 45.56 |
| RAMAN RE+ Comp +NIR RE | 55.98 |
| RAMAN RE+ Comp + RAMAN TR | 51.86 |
| RAMAN RE+NIR RE+ RAMAN TR | 47.94 |
| Comp+ NIR RE + RAMAN TR | 48.52 |

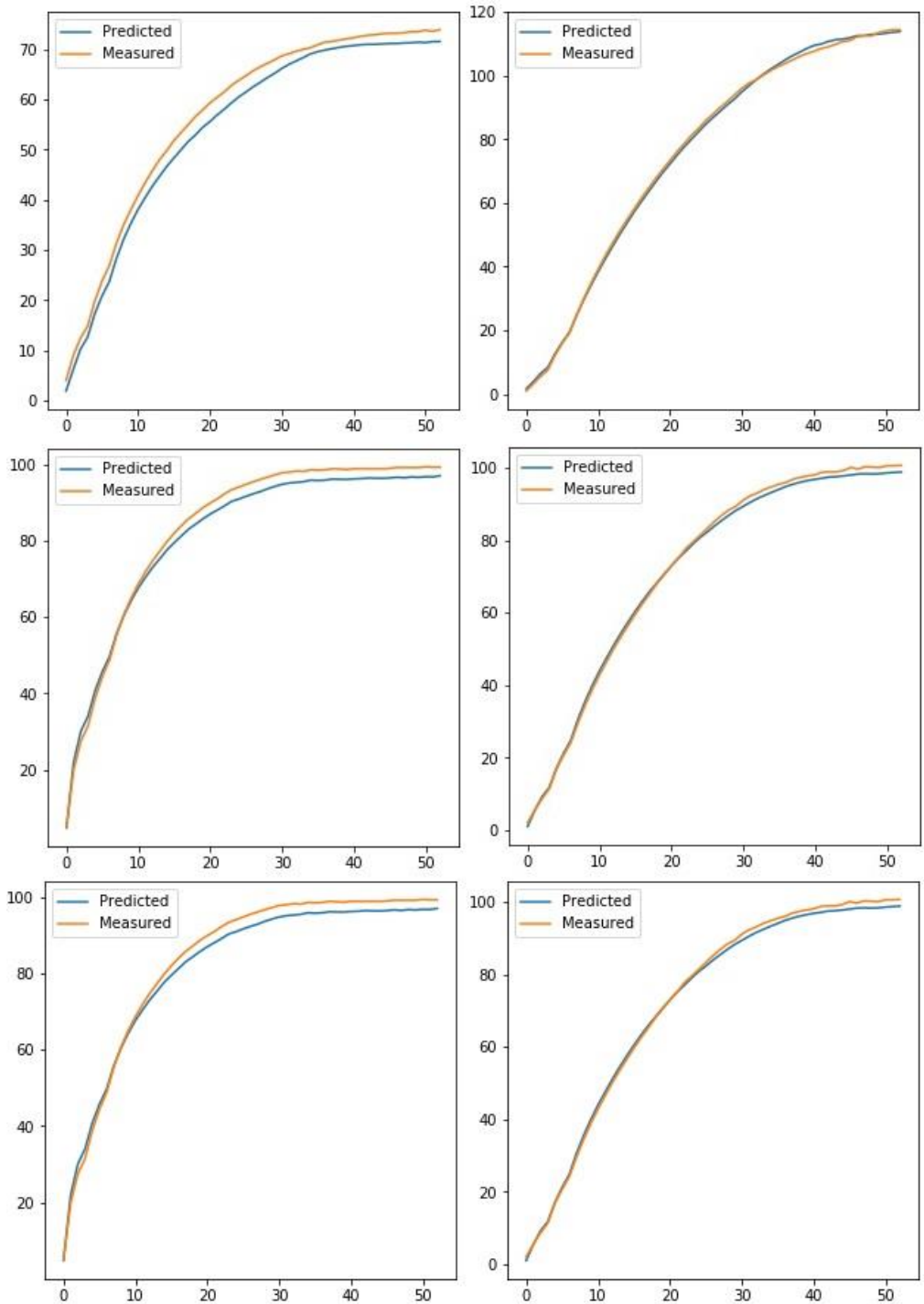

Figure 5. Sample predicted dissolution curves using NIR TR+ Comp+ NIR RE combination

## 4. CONCLUSIONS

The current work aimed to compare the measurements in the prediction of dissolution profiles using artificial neural network models. The spectroscopy data along with the compression force were standardized, and their dimensionality were reduced using PCA. ANN models were created using these data as input both as individual measurements, then a combination of two measurements then finally three measurements. The results showed that using only the NIR transmission method along with the compression force data or the Raman and NIR reflection methods, the dissolution profile was estimated within the acceptance limits of the $f_2$ similarity factor. The results showed that NIR transmission and the compression force are very important in the prediction of dissolution profiles, adding further measurements to these two slightly improved the results.

**ACKNOWLEDGEMENTS**

Project no. FIEK_16-1-2016-0007 has been implemented with the support provided from the National Research, Development and Innovation Fund of Hungary, financed under the Centre for Higher Education and Industrial Cooperation Research infrastructure development (FIEK_16) funding scheme.